# Prompt-to-Prompt: Text-Based Image Editing Via Cross-Attention Mechanisms - The Research of Hyperparameters and Novel Mechanisms to Enhance Existing Frameworks

Linn Bieske[1], Carla Lorente[1]

[1] Massachusetts Institute of Technology, Department of Electrical Engineering & Computer Science

## Abstract

Recent advances in image editing have shifted from manual pixel manipulation to employing deep learning methods like stable diffusion models, which now leverage cross-attention mechanisms for text-driven control. This transition has simplified the editing process but also introduced variability in results, such as inconsistent hair color changes. Our research aims to enhance the precision and reliability of prompt-to-prompt image editing frameworks by exploring and optimizing hyperparameters. We present a comprehensive study of the "word swap" method, develop an "attention re-weight method" for better adaptability, and propose the "CL P2P" framework to address existing limitations like cycle inconsistency. This work contributes to understanding and improving the interaction between hyperparameter settings and the architectural choices of neural network models, specifically their attention mechanisms, which significantly influence the composition and quality of the generated images.

## 1. Introduction

Recent advancements in image editing have transitioned from traditional pixel manipulation to the use of deep learning techniques, such as stable diffusion models. These methods have evolved from requiring users to manually mark areas for editing, as discussed by (Nichol et al., 2021), (Ramesh et al., 2022), and (Avrahami et al., 2022a), to the more advanced approach of using cross-attention mechanisms for text-driven control, introduced by (Hertz et al., 2022). This evolution represents a significant leap in simplifying the editing process.

However, the performance of these prompt-to-prompt editing frameworks, while promising, exhibits variability that impacts user expectations, such as inconsistent results in tasks like changing hair color (see Figure 1). This is due to the underlying stochastic nature of stable diffusion models. Our research focuses on benchmarking prompt-to-prompt frameworks, exploring the influence of its hyperparameters, and identifying enhancements to improve its performance and reliability in practical applications.

Building upon the code accompanying the publication by (Hertz et al., 2022), our research will introduce significant contributions to the field of image editing. (1) We present a comprehensive hyperparameter study of the "word swap" prompt-to-prompt editing method to identify optimal settings for various editing tasks. (2) Furthermore, we aim to generalize these optimized hyperparameters through the development of an "attention re-weight method," enhancing the adaptability of the editing framework. (3) Lastly, we propose a new framework designed to address the limitations observed in current methodologies such as cycle inconsistency, promising more consistent and reliable image editing outcomes.

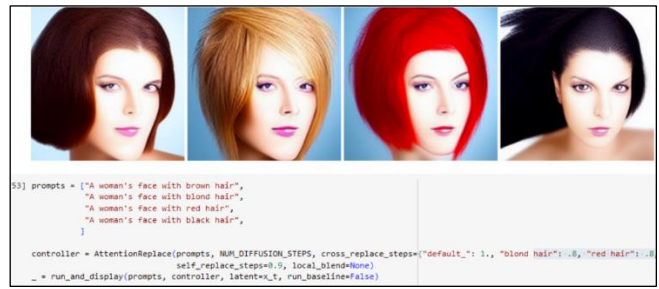

*Figure 1.* Stable diffusion models can generate significantly different images even if only one word in the submitted prompt is changed (underlying algorithm presented by Hertz et al., 2022). This is a shortcoming for prompt-to-prompt image editing.

Overall, our goal is to create frameworks that allow the precise iteration and editing of images generated with stable diffusion models using text prompts only.

## 2. Related Work

The evolution of image editing techniques using diffusion models has facilitated a transformative shift from traditional approaches to more sophisticated, AI-driven methods. Prior to the introduction of the prompt-to-prompt editing method, several significant methodologies emerged, as outlined below.

**Adding Noise and Denoising with a Prompt:** One foundational technique is by (Meng et al., 2021). In this approach, an image is first corrupted with noise and then denoised using a user-provided prompt, which specifies the desired outcome using pixel patches from reference images. This method allows for flexible yet precise control over the editing output. Similarly, (Liew et al., 2022) employs a pre-trained text-to-image diffusion model to blend and modify images based on extracted semantics. Here, the process involves crafting layout noises from an initial image and progressively mixing two distinct concepts through denoising, as guided by text describing the content semantics. This method maintains layout integrity while integrating new semantic elements.

**Noise Addition, Denoising with Prompts, and Mask Usage:** (Avrahami et al., 2022b) introduces a method where noise is added to an image, and a text-driven prompt guides the denoising of only masked regions, leaving the unmasked areas untouched. This process relies on predefined masks to specify editing zones. A more advanced iteration is presented by (Luddecke and Ecker, 2022) and (Couairon et al., 2022), where segmentation models or the diffusion process itself generates the masks, reducing the manual effort required

**Fine-tuning on a Single Image:** A different approach involves fine-tuning diffusion models on single images to achieve specific edits, as explored in by (Kawar et al., 2023) and (Valevski et al., 2023). This method is resource-intensive but allows for highly customized edits by adapting the model closely to the target image.

**Introduction to Prompt-to-Prompt Editing:** Building on these foundational methods, the prompt-to-prompt editing technique represents a significant leap forward. It simplifies the image editing process by eliminating the need for manual mask creation and the intensive computational resources required for fine-tuning. This method leverages the intuitive power of natural language, allowing users to specify edits directly through text prompts. The simplicity, flexibility, and user-friendliness of the prompt-to-prompt method offer a stark contrast to the complexities associated with previous techniques (Hertz et al, 2022).

In our research, we will further explore the capabilities and refine the applications of the prompt-to-prompt method, aiming to enhance its performance and user experience.

## 3. Method

In this study, we have constructed a robust machine learning pipeline using a Hugging Face stable diffusion model, which is specifically configured to allow access and manipulation of attention layers. This setup enables the development and refinement of editing prompts, the tuning of hyperparameters, and the systematic evaluation of the generated images. This comprehensive approach ensures that each component of the pipeline contributes effectively to the overall performance and accuracy of the image editing process.

### 3.1 Hyperparameter Study of Prompt-to-Prompt Editing Method "Word Swap"

This subsection analyzes the hyperparameters involved in the "word swap" method within the prompt-to-prompt editing framework, where "word swap" entails changing just one word in a prompt while keeping the rest unchanged. Drawing on foundational studies, particularly by Hertz et al. (2022), we have set default values for key hyperparameters and are systematically testing their impact on editing through empirical experiments.

Our study primarily adjusts hyperparameters like the silhouette threshold ('k'), cross-attention injection ('cross replace steps'), and self-attention injection ('self replace steps') to gauge their effects on the fidelity and precision of image edits. The silhouette threshold ('k') sets the editable areas of an image by converting attention masks into a binary filter matrix, focusing edits on specific regions. Cross-attention injection modulates how much the reference image influences the target image during diffusion, aligning edits with the intended style. In contrast, self-attention injection controls how much of the target's original attributes are preserved, enhancing authenticity and coherence in the edits. We also evaluate cycle consistency to ensure the method's reliability and the non-destructiveness of edits. Optimizing these hyperparameters is crucial for enhancing the effectiveness and adaptability of the "word swap" method in practical applications.

### 3.2 Generalization of Optimized Hyperparameters to "Attention Re-Weight Method"

Upon establishing optimal settings for the hyperparameters, we extend our analysis to their application in the attention re-weight method. Attention re-weighting in image editing is a technique that adjusts the focus of the model's attention on specific words within a prompt as proposed by (Hertz et al., 2022). This part of our study utilizes visual tools, such as comparative GIFs, to demonstrate the effects of hyperparameter adjustments on image generation. Specifically, we compare outcomes under two sets of conditions using the prompt "A woman's face with long wavy blond hair."

### 3.3 Proposal of New Framework

The traditional settings, particularly related to hyperparameters such as silhouette threshold, cross-attention, and self-attention injection, often result in overly constrained image geometries and poor utilization of self-attention mechanisms, leading to precision issues in the resultant edits. Moreover, the existing approaches lack cycle consistency, which is crucial for ensuring that edits can be reversed or iterated upon effectively. To address these challenges, we propose a novel framework, the "CL P2P" prompt-to-prompt image editing framework.

## 4. Results
### 4.1 Hyperparameter Study of Prompt-to-Prompt Editing Method "Word Swap"

Our research delved deeply into the hyperparameters that significantly influence the performance of the "word swap" prompt-to-prompt editing method.

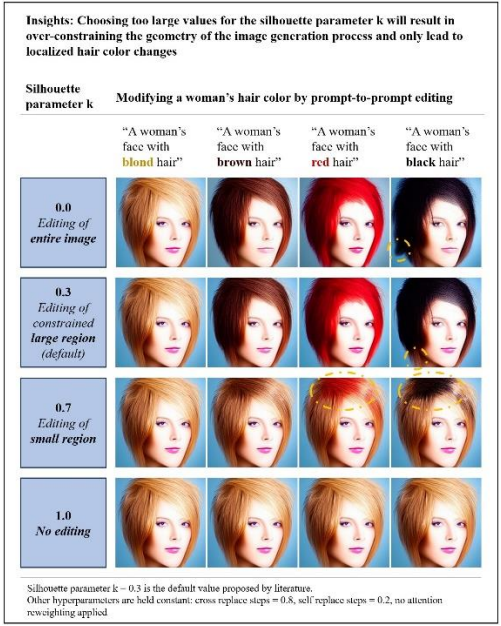

*Figure 2.* Impact of varying the silhouette parameter k on hair color editing using prompt-to-prompt image generation. Rows display k values from 0.0 (full image editing) to 1.0 (no editing), demonstrating how editable region size affects hair color precision and localization across prompts for blond, brown, red, and black hair.

For the **silhouette threshold hyperparameter** (k), we hypothesized that smaller values would allow broader areas of both the reference and target image to influence the final output, leading to potentially higher variability in the target images. Despite expectations, even at small values, the generated target image showed high similarity to the reference image, although some undesired geometric deviations remained, highlighting the importance of the self-attention mechanism. Conversely, larger k values resulted in excessive constraints on image geometry, limiting changes to very localized areas like color modifications in the reference image. We found that the optimal k value depends on the editing context, with a range of 0.0 to 0.3 suitable for hairstyles and 0.0 to 0.4 for landscapes (Figure 2, Appendix).

In terms of **cross-attention injection**, our hypothesis was that shorter injection durations would lead to significant geometric deviations between the target and reference images due to insufficient guidance. Contrary to this, even minimal cross-attention steps (e.g., 0.01 or 0.2) produced high-quality images when combined with a higher number of self-attention steps, indicating that minimal cross-attention is necessary for high-quality editing. Longer durations, though theoretically beneficial for geometric adaptation, were found to overly constrain the images, inhibiting perfect matches with the reference geometry. Thus, we recommend a lower number of cross-attention steps, as larger durations tend to over-constrain (Figure 3).

For **self-attention injection**, we anticipated that minimal durations would fail to achieve desired similarities between the reference and target images. Our findings challenge existing literature, showing that higher self-attention injection values actually lead to better geometric adaptation and similarity between the edited and target images. Therefore, a high value of self-attention injections should be prioritized over cross-attention injections, with the optimal value being context and task-dependent, recommended at 1.0 for hairstyle editing and 0.6 for landscapes (Figure 4).

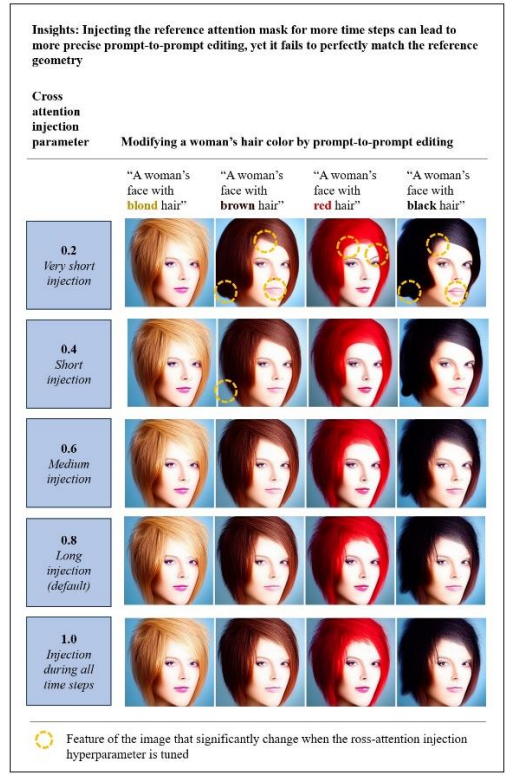

*Figure 3.* To explore the isolated impact of the cross-attention injection hyperparameter we switched off the localization of the editing (silhouette parameter k = 0). Consequently, the entire image can be edited. The self-attention parameter was held constant at 0.2. Cross replace steps is the default value proposed by the literature (Hertz et al., 2022).

**Conclusion hyperparameter study: Self-attention is all you need.** In contrast to the two other hyperparameters studied (silhouette parameter k and cross-attention injection

"cross replace steps", the self-attention hyperparameter ("self replace steps") is the one that shows the strongest impact on the geometric adaptability of the generated faces and hairstyles to match the reference image. This finding suggests that the underlying prompt-to-prompt image generation and editing mechanisms can be improved with regards to two aspects: (1) Precision of the editing algorithm which modifies variant features and shows stability for desired invariant features. (2) Computational efficiency of the underlying cross-attention algorithm e.g., by removing the constraints of the silhouette parameter.

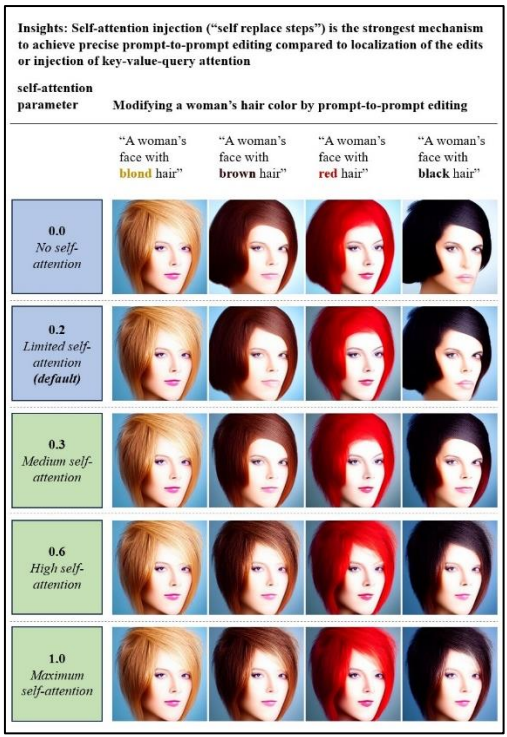

*Figure 4.* To explore the isolated impact of the self-attention hyperparameter we switched off the localization of the (silhouette parameter k = 0) and reducing the cross-attention injection steps to a minimum (0.2). Consequently, the entire image can be edited, and the number of cross-attention injection steps is held constant when the self-attention hyperparameter is tuned.

The optimal hyperparameter settings for the image editing method are: silhouette parameter k at 0.0 for maximum editing flexibility, cross replace steps at 0.2 for balanced reference influence, and self-replace steps at 0.8 to better retain the target image's characteristics, optimizing editing precision and adaptability.

Finally, our study addressed the **cycle consistency** of the method. We hypothesized that the method might lack cycle consistency due to limitations in how transformer models utilize reverse transformations. Our experiments confirmed this hypothesis, as the method failed to accurately reverse edits in complex transformations such as changing color shades from dark to light (Figure 5). This cycle inconsistency highlights a critical area for further refinement in ensuring the method's reliability and effectiveness in real-world applications.

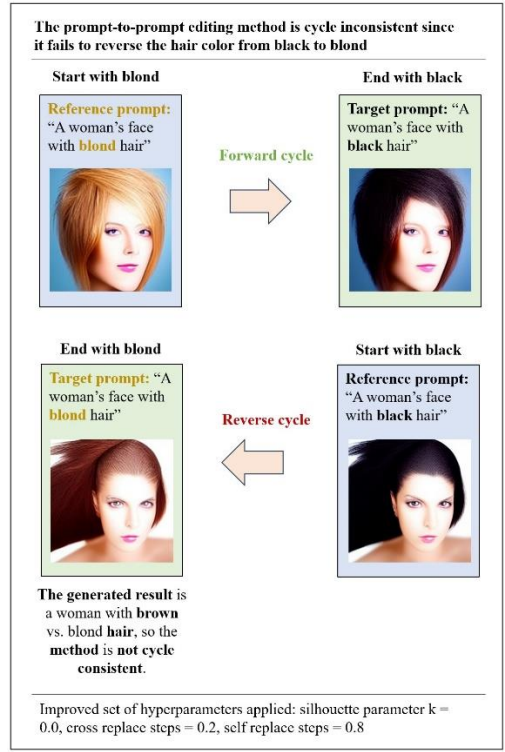

*Figure 5.* The studies prompt-to-prompt image editing algorithm is not cycle consistent. Reversing the prompt from black to blond does not generate an image of a blond woman.

### 4.2 Generalization of Optimized Hyperparameters to "Attention Re-Weight Method"

In the initial condition, images are generated using default parameters (k = 0.3; 'cross replace steps' = 0.8; 'self replace steps' = 0.2), where we observe certain instabilities and suboptimal results, particularly with negative weight assignments (Figure 6, left). Conversely, we refined hyperparameters which consistently yield more stable and accurate representations (Figure 6, right).

### 4.3 "CL P2P" prompt-to-prompt image editing framework

Optimization of critical hyperparameters significantly enhances the precision of prompt-to-prompt image editing, ensuring better similarity between reference and target images, particularly for faces and hairstyles. We recommend the following optimized values: (1) Silhouette parameter k: 0.0 to eliminate local editing, which hasn't shown significant precision improvements. (2) Cross-attention injection (cross replace steps): Reduced to 0.2 to increase geometric adaptability and improve alignment between reference and target images. (3) Self-attention injections (self-replace steps): Increased to 0.8 of the diffusion steps to maximize geometric adaptability, crucial for accurately editing hairstyles (Figure 7).

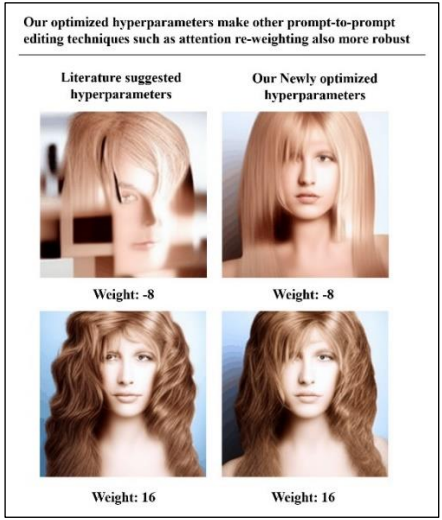

*Figure 6.* Prompt: "A woman's face with long wavy blond hair". Attention reweighting was applied to the word "wavy" such that both wavy and straight hair was generated. The current method using hyperparameters presented by (Hertz et al., 2022), fails to robustly generate straight hair. Whereases our optimized method has this capability.

To address cycle-inconsistency, we suggest balancing the asymmetry in the transformer model's usage of V values. Traditionally, the method only used the V values from the reference prompt, contributing to cycle-inconsistency. We propose introducing a new hyperparameter, "V value injection steps", to incorporate V values from both reference and target prompts, enhancing the model's cycle-consistency. This method allows for the precise control of injection steps, aligning more closely with the original and target features (Figure 8).

*Figure 7.* Comparison of traditional (Hertz et al., 2022) and 'CL P2P' hyperparameter settings and their impact on image editing. The table and images illustrate changes in local editing, cross-attention, and self-attention parameters, showcasing improved precision and consistency with the 'CL P2P' method.

*Figure 8.* Comparison of algorithmic steps between the traditional method (Hertz et al., 2022) and our 'CL P2P' method, highlighting enhancements in image editing and cycle consistency.

## 5. Discussion

This research contributes significantly to explaining the role of hyperparameter choices and transformer architecture decisions, particularly its attention mechanisms, in the stable diffusion model. It fosters a better theoretical understanding of how the composition of generated images is dependent on these underlying architectural choices and hyperparameters.

Overall, our research and development of the "CL P2P" framework has significantly enhanced the precision of prompt-to-prompt image editing by studying and optimizing critical hyperparameters, particularly increasing the impact of self-attention injection. This advancement has improved the alignment between generated and reference images and reduced the inherent stochasticity and variability of outcomes observed with similar prompts in current image generation models.

Looking forward, the "CL P2P" framework requires further development to enhance cycle-consistency. Future research should focus on optimizing "V value injection steps" to ensure the reversibility and integrity of the editing process. Additionally, expanding beyond singular reference and target prompts could transform user interactions. Integrating methods like "word swap," "attention re-weighting," and "prompt refinement" would enable dynamic, conversational interactions, allowing for continuous, iterative editing processes. This approach, not yet supported by existing generative models, promises to push the frontiers of image editing technology, enabling users to modify images through natural dialogue and interactive engagement, fostering interpretable and adaptable image generation.

**Acknowledgements**

We would like to thank Prof. Phillip Isola for his guidance and support. His expertise and insightful feedback have enriched our understanding and contributed to our findings.

Usage of generative AI: The writing of this paper was supported by ChatGPT to ensure a high quality of English.

**Appendix**

## 6. Underlying image generation and image editing mechanisms incorporated in stable diffusion model
### 6.1 Silhouette parameter

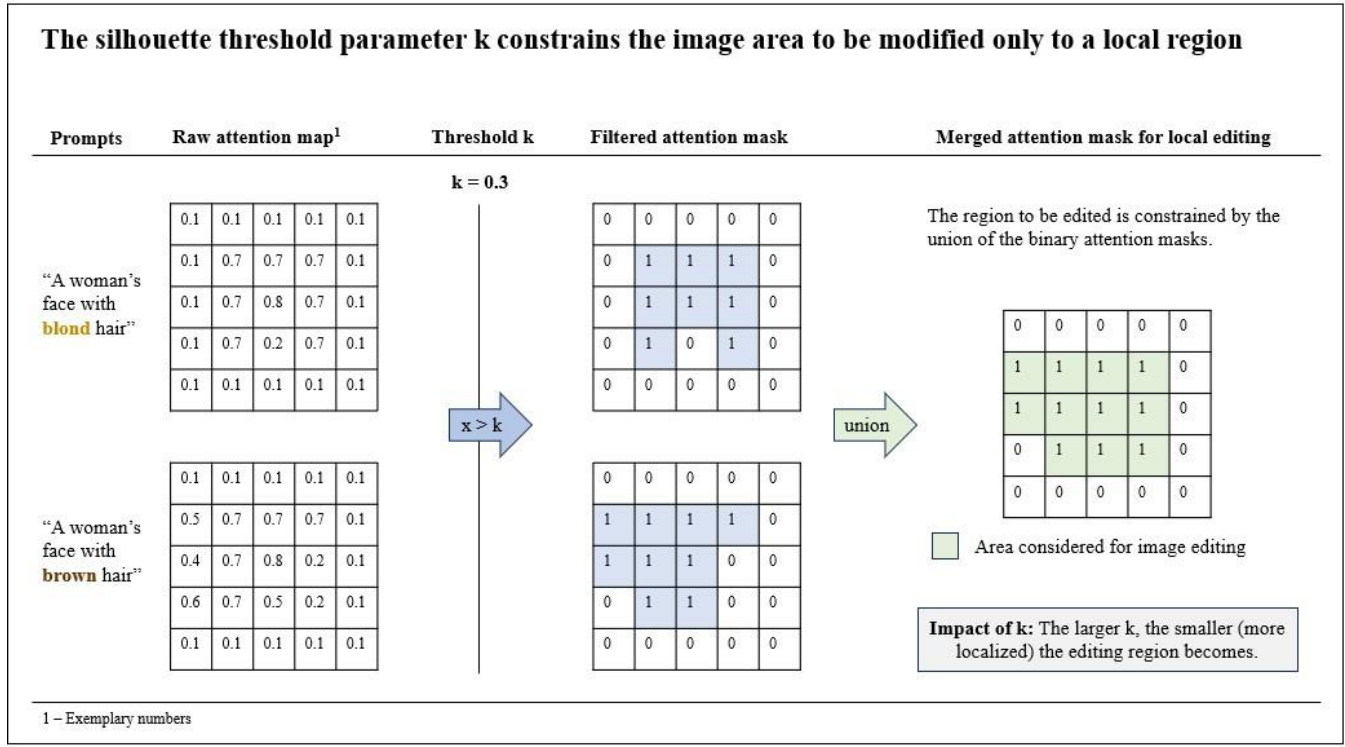

### 6.2 Cross replace steps (attention)

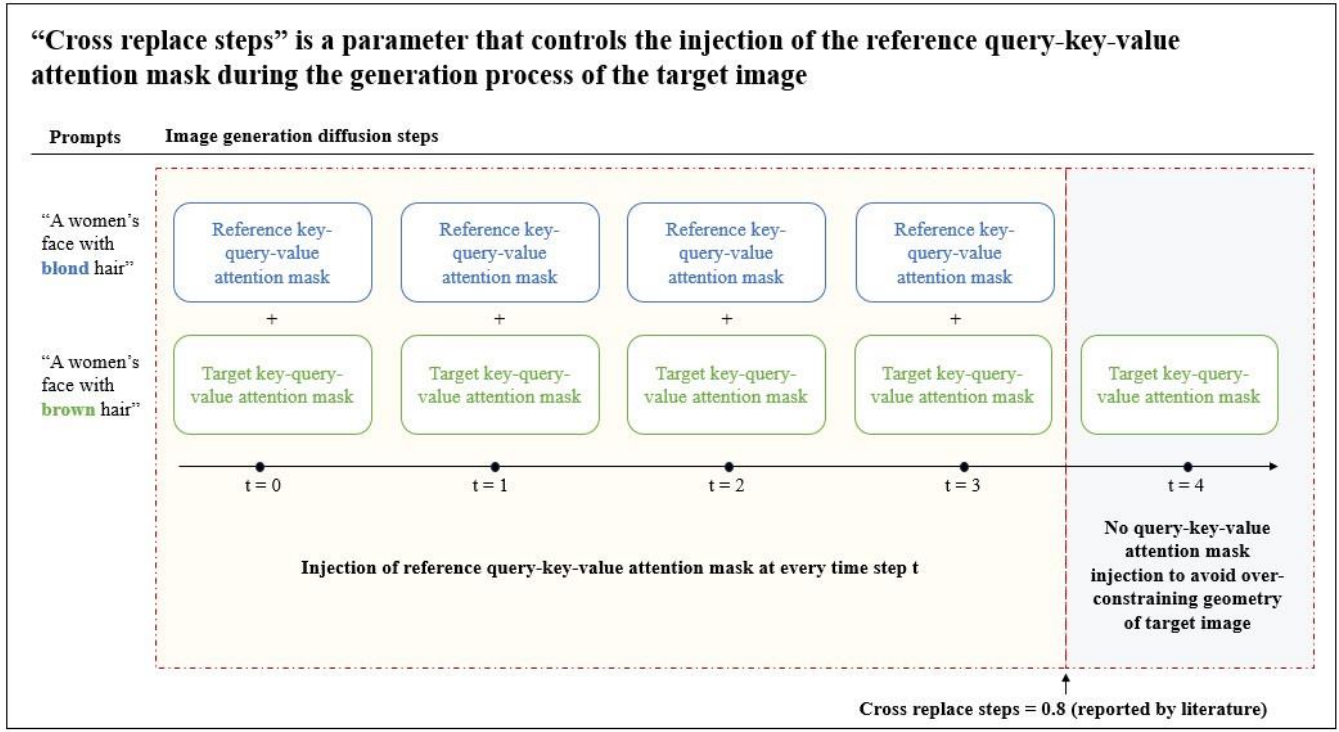

### 6.3 Self Replace Steps (Attention)

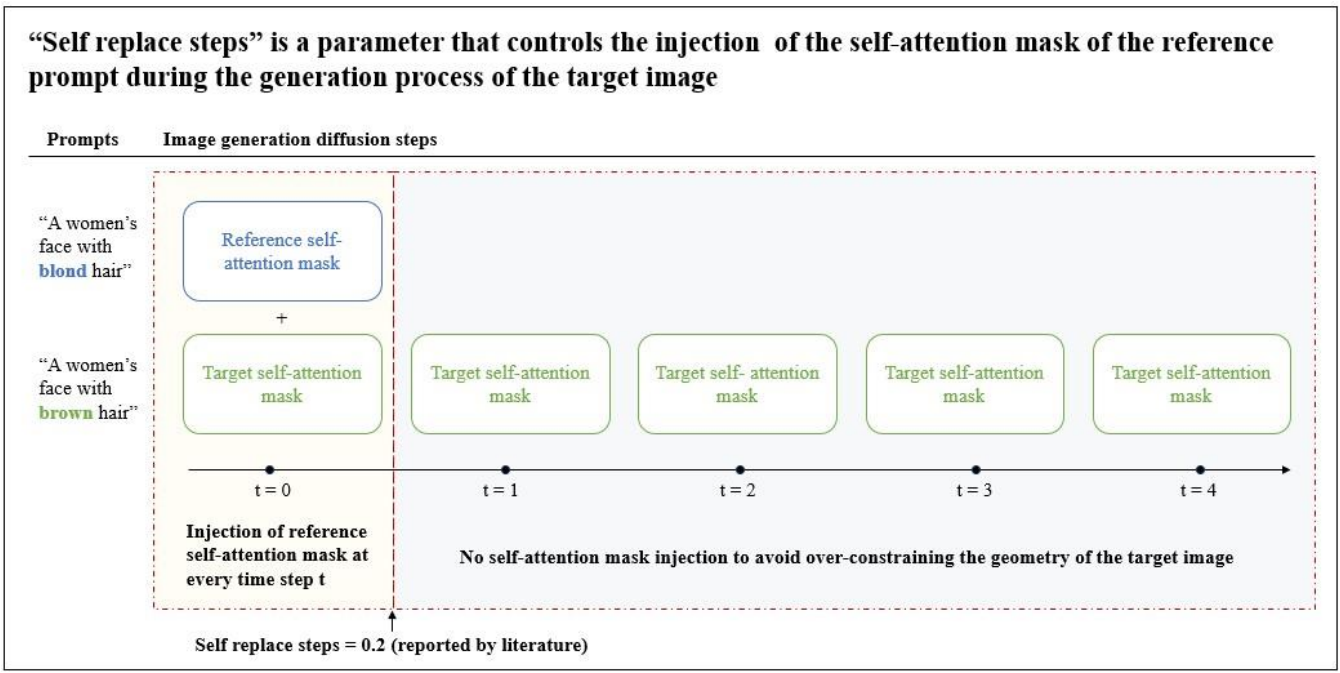

## 7. Additional Research Findings – Hyperparameter Study

**Insights: In dependence on the image domain and editing task, different silhouette threshold parameter k should be chosen to select larger or smaller areas to be edited**

| Silhouette parameter k | Modifying a landscape based on prompt-to-prompt editing | | | | Findings |
|---|---|---|---|---|---|
| | "A river between mountains" | "A street between mountains" | "A forest between mountains" | "A desert between mountains" | Only **one keyword** indicating the woman's hair color was **changed in** the **prompt** to edit the image. |
| **0.0** Editing of entire image | | | | | Even **without constraining** the region to be edited (k = 0), the generated **images** are of **high quality**. This indicates that the optimal value of the silhouette parameter k is context and task dependent. |
| **0.3** Editing of constrained large region (default) | | | | | When comparing the images for **k** values between 0 and 0.4 **almost no differences** in the generated images for a specific prompt can be observed. The desert of the first row looks identical to the desert of the second row. This indicates that there is **flexibility in choosing the k parameter for landscape images**. |
| **0.5** Editing of medium sized region | | | | | Once the **silhouette parameter is too high** (k = 0.5) the geometry of the image generation process is constrained too much such that e.g., **"flooding of the desert" starts**. Also, the **mountains** appear no longer in a "desert style" but **preserve the white color** of the original image. |
| **1.0** No editing | | | | | Setting the silhouette parameter to one (**k = 1**) will result in a total constraint **not leaving any remaining region** of the image to **be edited**. Therefore, **all four landscapes** in this row **are identical** and no image editing is performed. |

Silhouette parameter k = 0.3 is the default value proposed by literature.
Other hyperparameters are held constant: cross replace steps = 0.8, self replace steps = 0.2, no attention reweighting applied

## Insights: The optimal number of time steps for injecting the key-query-value attention mask of the reference image is context dependent, too large values over-constrain

To explore the isolated impact of the cross-attention injection hyperparameter ("cross replace steps") we switched off the localization of the editing by setting the silhouette parameter k = 0. Consequently, the entire image can be edited. The self-attention parameter ("self replace steps") was held constant at 0.2. Cross replace steps is the default value proposed by the literature.

| Cross attention injection parameter | Modifying a landscape based on prompt-to-prompt editing | Findings |
|---|---|---|
| | "A river between mountains"  "A street between mountains"  "A forest between mountains"  "A desert between mountains" | |
| 0.2 Very short injection | | The cross-attention injection mechanism allows tuning of distinct features of the landscape image: the **sky**, the **mountain range** and the **foreground** (street, forest, desert). |
| 0.4 Short injection | | Similar to the prompt-to-prompt editing of the women's faces with different hair colors, **significant deviations from the reference image** emerge for **short cross-attention injection times** (0.2 - 0.4). While the foreground features look very realistic, the mountains and the sky in the background show deviations. |
| 0.6 Medium injection | | Moreover, we conclude that there is **not one optimal hyperparameter value** for the cross-attention injection, but the user might decide what features are most important to them when evaluating the edited output image. An **acceptable range** could fall between **0.6 – 1.0**. |
| 0.8 Long injection (default) | | |
| 1.0 Injection during all time steps | | For example, for the prompt *"A forest between mountains"* a hyperparameter value of **0.6** would generate a **satisfying image** while the image generated for **0.8** also would look **acceptable**. |

Reference image remains unchanged | All three elements are changed by increasing cross-attention injection: the mountains, trees, and street width. | As the cross-attention injection is increased the number of trees, and their height is reduced. | With increasing cross-attention injection the mountain range looks more alike to the reference image.

○ Feature of the image that significantly change when the ross-attention injection hyperparameter is tuned

**Insights: The optimal value of the self-attention hyperparameter ("self replace steps") is context and task dependent – for the hair style editing 1.0 whereas for landscapes editing 0.6 is recommendable**

To explore **the isolated impact of the self-attention hyperparameter** ("self replace steps") we switched off the localization of the editing by setting the **silhouette parameter k = 0** and reducing the **cross-attention injection steps to a minimum** (**"cross replace steps" = 0.2**). Consequently, the **entire image can be edited**, and the **number of cross-attention injection steps is held constant** when the self-attention hyperparameter is tuned.

| self-attention parameter | Modifying a woman's hair color by prompt-to-prompt editing | Findings |
|---|---|---|
| | "A river between mountains" / "A street between mountains" / "A forest between mountains" / "A desert between mountains" | Only **one keyword** indicating the woman's hair color was **changed in** the prompt to edit the image. |
| **0.0** No self-attention | | Both the **mountains and the sky show strong deviations** from the geometry of the reference image. |
| **0.2** Limited self-attention (default) | | Of the **default value of 0.2**, the mountains still show **variation in their geometry compared to the reference image**. The similarity of the sky improves through the application of the self-attention mechanism. |
| **0.3** Medium self-attention | | For self-attention parameter **values greater than 0.3** lead to **geometric improvement** of both facial features (lips, noses) and hairstyle features (shape and texture of the hair). For a value **of 0.3**, the **lips still show deviations** for the women with brown or black hair. |
| **0.6** High self-attention | | **0.6 is the recommendable value** of the self-attention hyperparameter in the context of the editing of the **landscape images**. |
| **1.0** Maximum self-attention | | If the self-attention value is too high, then **the street, forest, and desert start looking like a river** in grey, green, or red respectively. It becomes clear that in contrast to the editing of face images, it is recommended to **limit the self-attention hyperparameters to lower values than 1.0 for landscape editing tasks**. |

**Conclusion: How much self-attention you need is context and task dependent!**
Still, the **self-attention hyperparameter** showed **the strongest impact on the geometric adaptability** when prompt-to-prompt editing landscape images. Yet, the optimal value for landscape images differs from the recommendable value for face images (0.6 vs. 0.1). In contrast to the fixed set of hyperparameters reported by the literature, it is recommendable to **adapt the hyperparameters context and task dependent** to enhance the precision of the prompt-to-prompt editing mechanism.

## 8. Additional research findings – cycle consistency and method stability

The current method is cycle inconsistent since it fails to reverse the hair color from black to blond, its stability is strongly dependent on the set of hyperparameters selected

| Parameter/prompts | Current values reported in literature | Intermediate 01 (no local editing) | Intermediate 02 (+ reduced attention injection) | Our optimized hyperparameter values |
|---|---|---|---|---|
| Silhouette parameter k | 0.3 | 0.0 | 0.0 | 0.0 |
| Cross replace steps | 0.8 | 0.8 | 0.2 | 0.2 |
| Self replace steps | 0.2 | 0.2 | 0.2 | 0.8 |

Forward cycle — Reference prompt: "A woman's face with blond hair" (left) → Target prompt: "A woman's face with black hair" (right)

Reverse cycle — Reference prompt: "A woman's face with black hair" (right) → Target prompt: "A woman's face with blond hair" (left)

- The **current methods fails to reverse the hair color** from black **to blond** when prompted. Consequently, it's **cycle inconsistent**.
- **Local editing** has **no positive impact** on the cycle consistency and therefore **could be removed to simplify** the prompt-to-prompt editing **method**.
- When **low** hyperparameter values are chosen for both the **attention injection and self attention**, then the model **result is unstable**.
- Our **newly proposed set** of hyperparameters sightly improves the output by **slightly brightening** the target **hair color. Cross replacement** values should be **low** and **self attention** values should by **high.**